\documentclass{sig-alternate-05-2015}
\usepackage{hyperref}
\begin{document}

\setcopyright{acmcopyright}
  



%


\title{Conversational Recommendation System with Unsupervised Learning}
%
%
%
%
%

\numberofauthors{2} 
%
\author{
%
%
\alignauthor
Yueming Sun, Yi Zhang\\
       \affaddr{University of California, Santa Cruz}\\
       \email{\{yueming,yiz\}@soe.ucsc.edu}
\alignauthor Yunfei Chen, Roger Jin\\
       \affaddr{Rulai Inc., Sunnyvale, California, USA}\\
       \email{\{yunfei,roger\}@rulai.io}
}

\CopyrightYear{2016} 
\setcopyright{rightsretained} 
\conferenceinfo{RecSys '16}{September 15-19, 2016, Boston , MA, USA} 
\isbn{978-1-4503-4035-9/16/09}
\doi{http://dx.doi.org/10.1145/2959100.2959114}

\maketitle
\begin{abstract}

We will demonstrate a conversational products recommendation agent. This system shows how we combine research in personalized recommendation systems with research in dialogue systems to build a virtual sales agent. Based on new deep learning technologies we developed, the virtual agent is capable of learning how to interact with users, how to answer user questions, what is the next question to ask, and what to recommend when chatting with a human user. 

Normally a descent conversational agent for a particular domain requires tens of thousands of hand labeled conversational data or hand written rules. This is a major barrier when launching a conversation agent for a new domain. We will explore and demonstrate the effectiveness of the learning solution even when there is no hand written rules or hand labeled training data. 

\end{abstract}

%
%


%
%

%
%



\section{Introduction}
Recommendation systems have achieved much commercial success and are becoming increasingly popular in a wide variety of online stores. Current recommendation systems provided ranked lists or images of items for a given user. Virtual sales agent with the capabilities of personalized recommendation have been imagined for years. With the ever increasing popularity of smart phones, mobile devices and virtual reality systems, this becomes a very important problem that could be a huge impact in the near future.

Research in recommendation systems field has enjoyed wide success, largely because the techniques are not restricted on a particular domain and can be adapted to many domains and business. To launch a recommendation system to a new domain, researchers and engineers normally only need to use the domain's existing user click or transaction data to train the recommendation model, without requiring too much manually labeled data. 

However, dialogue system technologies behind virtual agents, such as Apple's ``Siri'', are usually developed for a particular domain. The major elements of the agents include a language understanding component that converts user input (text or speech or multi-modal) into abstract representation, an internal state update component that updates the memory about the conversation, a policy component that determines the next system acts, and a natural language generation component for generating response to the user. Each component can be rule based, machine learning based or using a hybrid approach. For example, in order to predict user intention and update the memory, a learning based system usually learns from many hand labeled data points that look like this \cite{DEK1}: 

$$\textit{``I don't mind, but it should serve Japanese food."}$$ $$Label: \{act=inform, slots:[food=Japanese]\}$$ This is a major barrier when launching a dialogue system to a new domain/system, because much efforts are needed to either collect hundreds of thousands of hand labeled training data for machine learning algorithm to learn or hire people to write many rules. 

Due to the limitation of existing dialogue system technologies, how to build a personalized conversational recommendation agent that can be easily adapted to many different domains or systems is a big challenge. This demo will demonstrate how we tackle this challenge and our efforts on developing such a general purpose conversational recommendation system Rul.ai. This system learns from unlabeled user conversations and delayed user actions, thus it can be adapted to a new domain or business with descent performance.  A video of the system in action is available at \url{https://youtu.be/V_7G3wYt0mw}.

\section{System Implementation}
Our system has several major components, as shown in Figure 1. Most of the components are similar to a typical dialogue system, except that a personalized recommendation engine is added and used by the policy component to influence machine actions.

\begin{figure*}[!ht]
  \centering
\includegraphics[width=\textwidth, height=5cm]{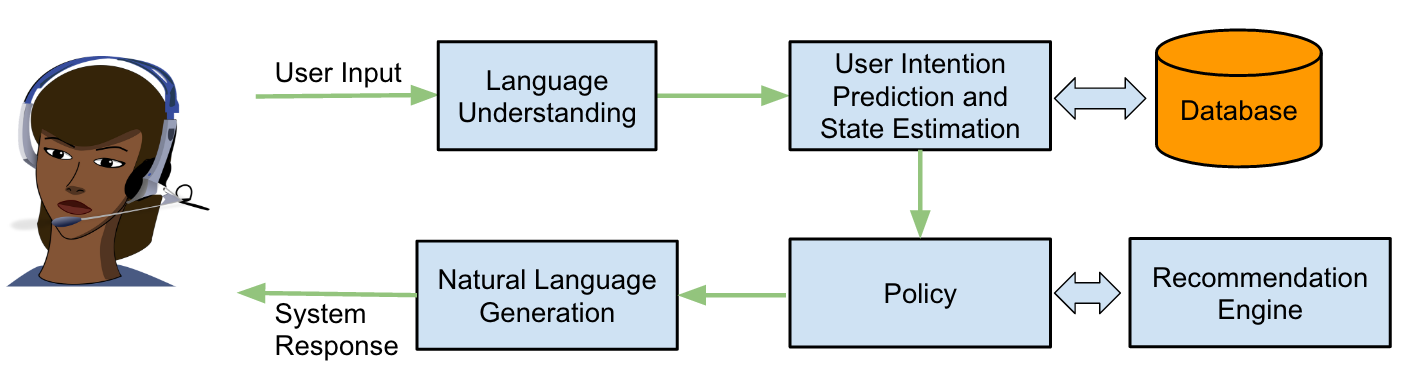}
\caption{System Architecture}
\end{figure*}

\subsection{User Intention Prediction and State Tracking}

The goal of a conversational recommendation agent is to finalize a purchasing order for the user at the end of the conversation. In order to achieve this goal, the users need to provide values for meta data needed for the agent to place the order. These information are also known as slot-value pairs \cite{DEK1}. For example, a pair $[type=latte]$ denotes the type of coffee ordered by user is ``latte''. 
After each user utterance, the user intention prediction and state tracking component needs to predict the current user's intention precisely and to update which state she/he is at. The state includes all slot-value pairs the agent has from the user so far, such as $[food=Japanese, location = 95070]$. Without requiring hand labeled data for each user utterance, we developed a new unsupervised learning method to train models used by this component. Our approach is to train a deep learning model \cite{DEK2} that learns from delayed rewards, which are the orders at the end of each conversation. These delayed rewards are collected by e-commercial companies in their normal business.
    
\subsection{Personalized recommendation and learning}

Recommendation in conversation is an open area for research. Besides user transaction history and user demographic information that are normally used in traditional recommendation engines, our engine also has a rich set of additional information about the user needs, such as possible user initial request (i.e. a user query) or supplemental information collected while talking with the user. To better serve users in our system, we have a recommendation system that's capable of using those information to rank products, generate recommendations, get user feedback to update memory and actively solicit additional user feedback.
    
\subsection{Machine utterance generation}

We will demonstrate two machine utterance generation mechanisms inside our system. One is a template based model with a statistical learning method to select a template and automatically fill the missing components in the selected template for each machine utterance generation request. The second solution is based on a deep learning model (i.e. a sequence to sequence language generation model). 

\section{Data}

We trained the model with unlabeled data collected by a mobile app, which provides personal assistance services to thousands of daily active users. The data set contains natural language chats between many users and real human agents (i.e. assistant). In a successful chat session, a real human agent guides a user through the processes to fulfill a purchasing order.

\section{A dialogue example}

\begin{figure}[!ht]\caption{Demo Screenshot}
\centering
\begin{minipage}{1.08\linewidth}
\centering
\includegraphics[width=\linewidth]{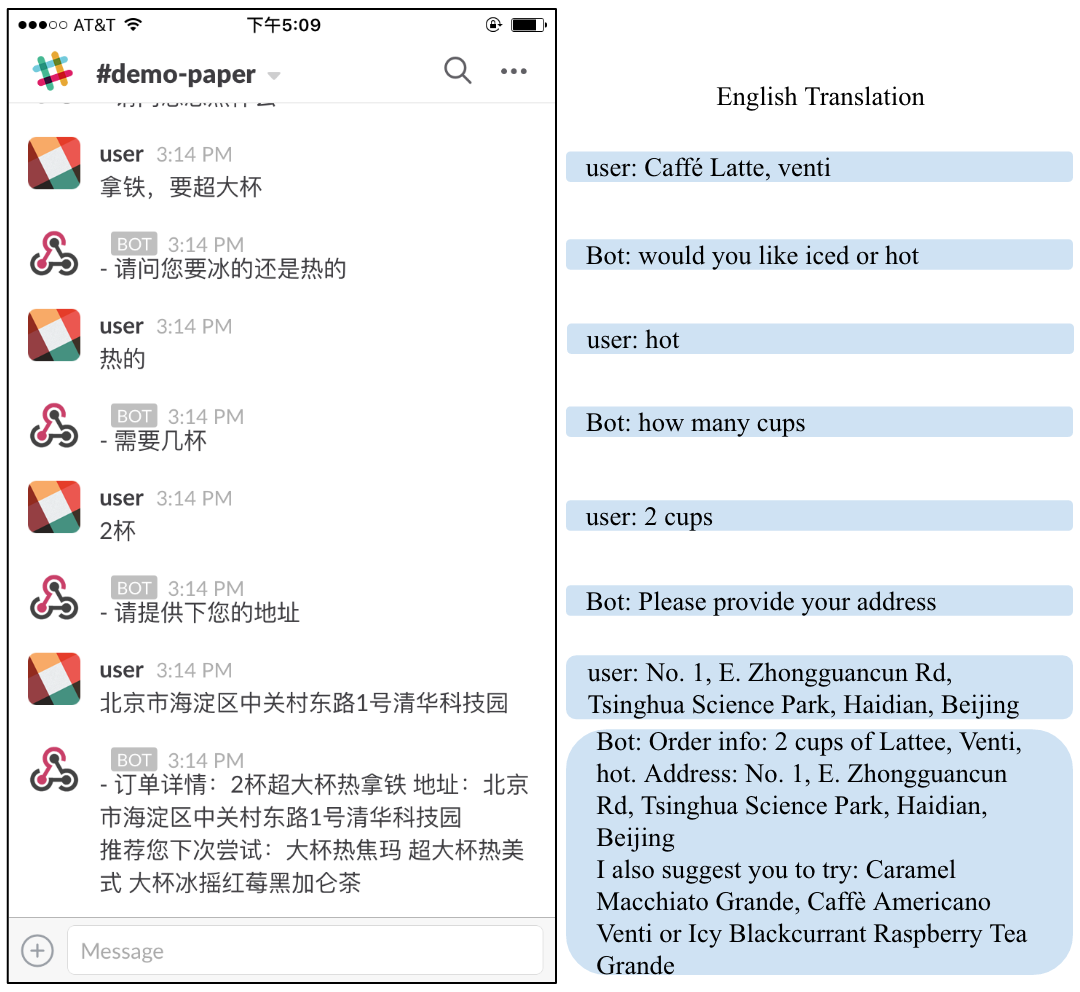}
\footnotesize
\newline
Both English and Chinese versions will be used for demo.
\end{minipage}
\end{figure}








\end{document}